\documentclass[runningheads,usenames,dvipsnames]{llncs}
\usepackage{graphicx}

\usepackage{paralist, tabularx} %
\usepackage{tabularx,booktabs}
\newcolumntype{Y}{>{\centering\arraybackslash}X}

\usepackage{amsmath,amsfonts,bm}

\def\1{\bm{1}}

\newcommand{\test}{\mathcal{D_{\mathrm{test}}}}

\def\vg{{\bm{g}}}
\def\vh{{\bm{h}}}

\def\vm{{\bm{m}}}

\def\vv{{\bm{v}}}

\def\vx{{\bm{x}}}

\def\vz{{\bm{z}}}

\DeclareMathAlphabet{\mathsfit}{\encodingdefault}{\sfdefault}{m}{sl}
\SetMathAlphabet{\mathsfit}{bold}{\encodingdefault}{\sfdefault}{bx}{n}

\newcommand{\x}{$\times$}

\usepackage{dsfont}
\usepackage{enumitem}
\usepackage{tikz}
\usepackage{comment}
\usepackage{amsmath,amssymb} 
\usepackage{color}
\usepackage{xspace}
\usepackage{xcolor}
\usepackage{nth}
\usepackage[colorlinks=true,citecolor=black]{hyperref}

\usepackage{colortbl}
\definecolor{knowsobj}{rgb}{1.0, 0.9, 0.9}  

\usepackage[numbers,sort]{natbib} 

\newcommand\knowsobj{\textcolor{red}{\pmb{yes}}}
\newcommand\doesntknowobj{\textcolor{lightgray}{no}}

\newcommand{\apbbox}[1]{AP$^\text{bb}$}
\newcommand{\apmask}[1]{AP$^\text{mk}$}

\usepackage{amssymb}
\usepackage{pifont}
\newcommand{\fcos}{FCOS$^\star$\@\xspace}

\begin{document}
\pagestyle{headings}
\mainmatter
\def\ECCVSubNumber{7509}  

\title{Object discovery and representation networks} 

\titlerunning{Object discovery and representation networks}
%
\author{Olivier J. Hénaff \and
Skanda Koppula  \and
Evan Shelhamer  \and \\
Daniel Zoran \and
Andrew Jaegle  \and 
Andrew Zisserman  \and \\
João Carreira \and 
Relja Arandjelović 
}
\authorrunning{O. Hénaff et al.}
%
\institute{DeepMind, London, UK}
\maketitle

\begin{abstract}

The promise of self-supervised learning (SSL) is to leverage large amounts of unlabeled data to solve complex tasks. While there has been excellent progress with simple, image-level learning, recent methods have shown the advantage of including knowledge of image structure. However, by introducing hand-crafted image segmentations to define regions of interest, or specialized augmentation strategies, these methods sacrifice the simplicity and generality that makes SSL so powerful. Instead, we propose a self-supervised learning paradigm that discovers this image structure by itself. Our method, \textbf{Odin}, couples \textbf{o}bject \textbf{di}scovery and representation \textbf{n}etworks to discover meaningful image segmentations without any supervision. The resulting learning paradigm is simpler, less brittle, and more general, and achieves state-of-the-art transfer learning results for object detection and instance segmentation on COCO, and semantic segmentation on PASCAL and Cityscapes, while strongly surpassing supervised pre-training for video segmentation on DAVIS. 

\end{abstract}

\section{Introduction}

Self-supervised learning proposes to leverage large amounts of unlabeled data to solve complex visual tasks.
Early attempts hand-designed pretext tasks, which required some semantic understanding of images and their layout to solve \cite{doersch2015unsupervised,pathak2016context,zhang2016colorful,noroozi2016unsupervised}.
Contrastive learning departed from this tradition by radically simplifying the self-supervised protocol, in that the pretext task is specified by the data itself: representations must learn to distinguish a given example from the others in the dataset \cite{hadsell2006dimensionality,dosovitskiy2014discriminative,wu2018unsupervised,oord2018representation}.
Modern instances of the contrastive framework have proven to be very powerful, leading to strong performance on a variety of downstream tasks \cite{henaff2019data,he2019momentum,chen2020simple}.
More recent self-supervised methods have simplified the framework further, removing the need for negative samples \cite{grill2020bootstrap}, bespoke architectural components  \cite{chen2021exploring}, and learning dynamics \cite{zbontar2021barlow}, suggesting that increasingly domain-agnostic and data-driven methods might enable learning from ever-larger and more general sources of data.  

However, a parallel line of work has asked whether the current self-supervised paradigm---which maximizes the similarity of the same data-point under different views---is too simple.
By treating data-points as monolithic instances, these methods overlook the complexity of real-world data: natural scenes are composed of many objects, natural speech of multiple speakers, and natural videos of many scenes.
Ignoring this variability and encouraging models to represent different parts of an image in a similar manner risks dampening their selectivity for objects, their relationships, and layouts in real-world scenes. 
Indeed, several works have demonstrated the benefits of properly handling this variability when learning task-relevant representations \cite{henaff2021efficient, van2021unsupervised, wei2021aligning, tomasev2022pushing, ryali2021learning}. While such object- and segmentation-aware approaches have yielded impressive empirical gains, they have relied on more domain-specific prior knowledge to expose the structure of the data---for example by using hand-crafted segmentation algorithms \cite{henaff2021efficient}, or salience estimation networks trained on human annotations \cite{van2021unsupervised}---bounding how much they can learn from the data and what data they can be used on. 

In this work we ask whether this knowledge can instead be derived from the data itself. To do so, we propose to couple two learning processes: object discovery and object representation. We use object discovery 
to uncover the structure of individual data-points, allowing the self-supervised task to focus on learning invariant representations of object-level instances. In turn, we use the resulting object representations 
as features for unsupervised object discovery, which feeds back into the representation learning process. These object discovery and representation networks thus engage in a virtuous cycle of representation and segmentation quality: better representations lead to better segmentations, and vice versa. Crucially, we derive the unsupervised segmentations with no prior knowledge about image structure or content, using a simple k-means clustering of local features to partition each image. We thus open the possibility of applying the algorithm to different domains, modalities, and their combination. 

We make the following contributions: 
\textbf{1)} Our object discovery networks  uncover, in an entirely self-supervised manner and without any prior knowledge of image structure or segmentation, meaningful decompositions of real-world scenes.
\textbf{2)} Our object representation networks lead to state-of-the-art results in transfer learning to object detection and instance segmentation on COCO, and semantic segmentation on PASCAL and Cityscapes, surpassing prior works which exploit segmentation and saliency information, without requiring this prior knowledge. 
\textbf{3)} Our object representation networks seamlessly generalize to video understanding, surpassing supervised pre-training for video object segmentation on DAVIS. Finally, we test the resilience of our method by varying its essential components, and find it to be very robust, supporting further computational benefits. Together these results suggest that knowledge of scene structure, and the benefits it confers in representing objects, can---with the right learning paradigm---be extracted from the data itself. 

\section{Related Work}

\noindent \textbf{Pre-contrastive self-supervised learning: hand-designed tasks.}
Early self-supervised approaches focused on injecting human expertise and intuition into the design
of proxy tasks for pretraining. For example, to stimulate the network to learn object parts,
\cite{doersch2015unsupervised} designed the task of predicting the spatial arrangement between
local image patches. A rich collection of such intuitions and objectives was further developed,
ranging from pixel-wise reconstruction-based approaches, such as denoising \cite{vincent2008extracting}, inpainting \cite{pathak2016context}, colorization \cite{zhang2016colorful,larsson2017colorization}, and more \cite{donahue2016adversarial,zhang2017split},
to higher-level pretext tasks, such as predicting spatial layouts \cite{doersch2015unsupervised,nathan2018improvements,noroozi2016unsupervised}, orientation \cite{gidaris2018unsupervised}, egomotion \cite{agrawal2015learning}, and temporal ordering \cite{misra2016shuffle}.

\vspace{0.5em} \noindent \textbf{Contrastive learning and its variants.}
Instance discrimination \cite{dosovitskiy2014discriminative} has proven to be a very powerful pretext task
which, we argue, owes its superior performance to being minimally hand-designed and maximally data-driven.  
By minimizing a contrastive loss \cite{hadsell2006dimensionality,oord2018representation},
the similarity of a representation across different `views' of the same image is maximized,
while minimizing their similarity with distracting negative samples.
Multiple views of a single data-point can naturally be extracted from multimodal or multisensory data \cite{arandjelovic2017look,recasens2021broaden,korbar2018cooperative,miech19howto100m,sun2019videobert,owens2018audio}
while for a single image-only modality they are typically constructed via local and global cropping \cite{hjelm2018learning,bachman2019learning,oord2018representation,henaff2019data} or data-augmentation \cite{chen2020simple,doersch2017multi,dosovitskiy2014discriminative,he2019momentum,wu2018unsupervised}. Positive pairs then correspond to views of the same data point, while negatives are sampled views of different data-points (typically from the same mini-batch),
although the need for negative samples has recently been questioned \cite{chen2021exploring,grill2020bootstrap,zbontar2021barlow}.

\vspace{0.5em} \noindent \textbf{Baking prior knowledge back into self-supervised learning.} A growing body of research has brought hand-designed supervisory signals back into the self-supervised paradigm. For example, \cite{henaff2021efficient, wei2021aligning, zhang2021looking, xie2021unsupervised, van2021unsupervised} decompose input images into their constituent objects and regions of interest using supervised segmentation algorithms, or hand-crafted heuristics. Object-level features are then computed for each region, and optimized using a contrastive objective. Other approaches use object-agnostic learning objectives, but integrate knowledge from segmentation heuristics or models in their augmentation strategies \cite{tomasev2022pushing, zhao2020distilling, mishra2021object}.  

This trend is reflected in the broader research in self-supervised learning for other modalities. For example, \cite{lin2021entitybert} uses domain-specific knowledge to improve the masking strategies of BERT and other masked-language models. \cite{charig2021self, han2020self} leverage motion and flow information to improve learning from video. And similar to previously described works in vision, \cite{nunes2022segcontrast} uses a segmentation step prior to applying SSL on point clouds. In all cases, we aim to remove the dependency on such prior knowledge while retaining its benefits for representation learning. 

\vspace{0.5em} \noindent \textbf{Clustering and representation learning.}
In parallel to the advent of contrastive methods, clustering-based representation learning methods have seen similar success, particularly in harnessing large amounts of uncurated images for transfer learning \cite{caron2018deep,asano2020self,caron2019unsupervised,caron2020unsupervised,goyal2021self,ji2019invariant}. Although they differ in their formulation of the self-supervised objective, these works also treat entire images as monolithic entities. 

In contrast, IIC \cite{ji2019invariant} performs within-image clustering using similarity losses counterbalanced by information maximization, obtaining compelling results in unsupervised segmentation. PiCIE \cite{cho2021picie} improves on this approach by imposing carefully-chosen, modality-specific geometric data augmentations and corresponding invariance and equivariance constraints. Neither of these works explicitly leverage their unsupervised segmentations for transfer learning across datasets and tasks however, which we investigate here.

\vspace{1em} \noindent \textbf{Object discovery.}
Recent years have seen a growing interest in developing generative models that perform object discovery. By introducing different inductive biases such as mixture-model likelihoods \cite{monet2019, greff2019multi}, attention \cite{locatello2020object, zoran2021parts} and specific forms of factorization \cite{kipf2021conditional, kabra2021simone}, such models are able to discover objects and their interactions \cite{goyal2019recurrent, didolkar2021neural, shanahan2020explicitly}. While much progress has been made, models from this family have yet to be demonstrated to work on natural images \cite{greff2019multi} and their application has been limited to synthetic data and simple highly structured environments typically used in robotics. Here we investigate object discovery on natural images in the wild, leveraging contrastive representation learning to enable this with simple k-means clustering. 

\section{Method}
\subsection{Self-supervised learning with Odin}

\begin{figure}[t]
\centering
\includegraphics[width=\textwidth]{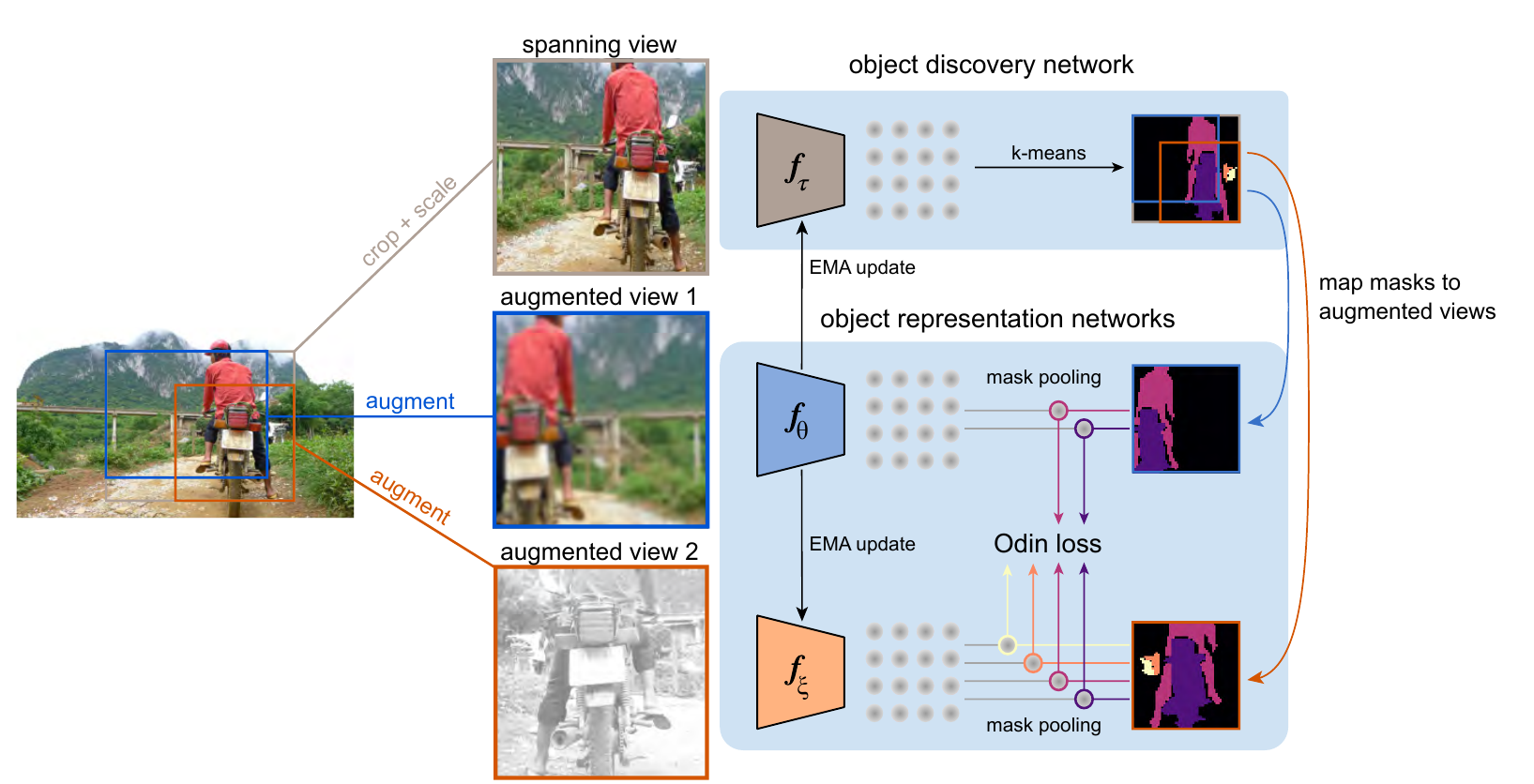}
\caption{Object discovery and representation networks. The object discovery network takes as input a cropped but otherwise un-augmented view of the image, and parses it using k-means clustering on its representation of it. The resulting segmentation is mapped into two augmented views of the same image, such that the masks are aligned across views and with the underlying image. The object representation networks take as input the augmented views of the image, and are trained using a self-supervised objective based on features pooled within each mask. The object discovery network is regularly updated with the parameters of the object representation network.} 
\label{fig:method}
\end{figure}

Our method learns two sets of networks which work in collaboration. The \textit{object discovery} network produces feature maps from high-resolution images. These feature maps are then spatially clustered to produce a segmentation of the image. The \textit{object representation} networks learns better features via a contrastive loss which uses the masks proposed by the object discovery network. The resulting improved features are then used by the object discovery network to create better segmentations, and this process is continuously repeated. Figure \ref{fig:method} illustrates the full method, which we detail below.

\vspace{1em} \noindent \textbf{Object discovery network: from representations to segmentations.} \\ Given an image $\vx$, we compute a spanning view $\vv^0$ which encompasses most of the area of the image (Figure \ref{fig:method}, \textit{spanning view}, defined below) and which is simply cropped and resized. We use a feature extractor $f_\tau$ to encode this view into a spatial map of hidden vectors $\vh^0 = f_\tau(\vv^0)$ and projections $\vz^0 = g_\tau(\vh^0)$, where $g_\tau$ is a two-layer MLP which is applied to each vector independently, and $\tau$ are the parameters of the \textit{object discovery network}. We apply $K$-means clustering to the spatial map of features $\vh^0$ or $\vz^0$, segmenting it (independently across images) into $K$ non-overlapping binary masks $\vm^{k,0}$ (Figure \ref{fig:method}, top row).

\vspace{1em} \noindent \textbf{Object representation networks: from segmentations to representations.} We produce two views $\vv^1$ and $\vv^2$ of the image by augmenting $\vx$ twice, using the random preprocessing pipeline of BYOL \cite{grill2020bootstrap}, which includes random cropping, flipping, blurring, and point-wise color transformations (Figure \ref{fig:method}, \textit{augmented views} and appendix).  

The \textit{spanning view} $\vv^0$ is chosen as the smallest crop which spans the spatial extent of the augmented views $\vv^1$ and $\vv^2$. We can therefore obtain two sets of masks $\vm^{k,1}, \vm^{k,2}$ which are consistent with each other and aligned with the underlying image content, by simply cropping, flipping, and resizing each mask $\vm^{k,0}$ as necessary (Figure \ref{fig:method}, right). Despite the significant differences in appearance across views, these masks contain the same underlying image content (up to differences in cropping), which we leverage in our objective. 

Each augmented view $\vv^l \in \{\vv^1, \vv^2\}$ is encoded with a feature extractor $f_\theta$ into a spatial map of hidden vectors: $\vh^l_\theta = f_\theta(\vv^l)$ where $\theta$ are the parameters of the \textit{object representation network} being optimized.  
For every mask $\vm^{k,l}$ in the image, we compute a mask-pooled hidden vector 
\begin{equation}
\vh^{k,l}_\theta = \frac{1}{\sum_{i,j} \vm^{k,l}[i,j] } \sum_{i,j} \vm^{k,l}[i,j] \ \vh_\theta^l[i,j],
\end{equation}
discarding masks that are empty (due to cropping). Our goal is to ensure that these object-level features are roughly invariant across views. Specifically, we wish for an object-level feature in one view to be predictive of the same image content in the other view. To that end we transform the object-level hidden vectors $\vh^{k,l}_\theta$ with two-layer MLPs $g_\theta$ and $q_\theta$, yielding non-linear \textit{projections} $\vz_\theta^{k,l} = g_\theta( \vh_\theta^{k,l} )$ 
and \textit{predictions} $q_\theta( \vz_\theta^{k,l} )$. 
In theory, we could regress the prediction $q_\theta( \vz_\theta^{k,1} )$ directly onto its target $\vz_\theta^{k,2}$, however it is helpful to stabilize these targets by encoding them instead with specific \textit{target networks} 
$g_\xi$ and $f_\xi$, where the parameters $\xi$ vary more slowly  \cite{grill2020bootstrap,tian2021divide,chen2021empirical,henaff2021efficient}. We therefore instead use the projections $\vz_\xi^{k,l} = g_\xi( \vh_\xi^{k,l} )$ as targets for the online prediction networks. 

\vspace{1em} \noindent \textbf{Jointly learning to discover and represent objects.}
Given a set of masks which approximately segment an image into objects, we wish to learn representations which distinguish these objects, while being invariant to identity-preserving transformations. Contrastive learning provides a straightforward objective for achieving this. Specifically, \textit{contrastive detection} \cite{henaff2021efficient} trains a network to recognize object-level features across views, in the presence of many distracting ``negative'' features from other objects. The resulting objective maximizes the similarity between different views of the same object, while minimizing the similarity between different objects. We define the similarity between object-level features across views as 
\begin{equation}
s_k^{1 \rightarrow 2} = \frac{1}{\alpha} \frac{\langle q_\theta( \vz_\theta^{k,1}), \vz_\xi^{k,2} \rangle}{ \lVert q_\theta( \vz_\theta^{k,1}) \rVert \lVert \vz_\xi^{k,2} \rVert}
\end{equation}
where $\alpha$ is temperature hyper-parameter. We define the similarity between an object-level feature and a distracting negative sample $s_k^{1 \rightarrow n}$ analogously, by replacing the paired feature $\vz_\xi^{k,2}$ with one from a different mask in the same image, or a different image altogether. The contrastive loss function for an individual feature is then  
\begin{equation}
\ell_k^{1 \rightarrow 2}(\theta; \xi, \tau) = - \log \frac{\exp( s_k^{1 \rightarrow 2} ) }{\exp( s_k^{1 \rightarrow 2} ) + \sum_n \exp( s_k^{1 \rightarrow n} )},
\end{equation}
which we sum across objects, views, and images in the mini-batch (summation across images not shown for clarity)
\begin{equation}
\mathcal{L}(\theta; \xi, \tau) = \frac{1}{K} \sum_{k=1}^K \ell_k^{1 \rightarrow 2}(\theta; \xi, \tau) + \ell_k^{2 \rightarrow 1}(\theta; \xi, \tau).
\end{equation}
We optimize the object discovery and representation networks using a strategy inspired by BYOL \cite{grill2020bootstrap}. 
One object representation network (the \textit{online network} with parameters $\theta$) is updated with gradients from the contrastive loss. The second object representation network (the \textit{target network} with parameters $\xi$) and the object discovery network (with parameters $\tau$) are updated using an exponential moving average of the online network:
\begin{align}
    \theta & \leftarrow \textrm{optimizer}(\theta, \nabla_\theta \mathcal{L}(\theta; \xi, \tau), \lambda_\theta) \\ 
    \xi & \leftarrow (1 - \lambda_\xi) \xi + \lambda_\xi \theta \\ 
    \tau & \leftarrow (1 - \lambda_\tau) \tau + \lambda_\tau \theta,
\end{align}
where the optimizer is LARS \cite{you2017large}, and $\lambda_\theta$, $\lambda_\xi$,  $\lambda_\tau$ are learning rates for the online, target, and discovery networks respectively.
We adopt the learning rates for online and targets networks from BYOL without modification.
For the object discovery network, we consider two schedules: a constant learning rate which continuously updates the object discovery network with the online one (e.g.\ $\lambda_\tau = 10^{-3}$), and a small number of discrete updates which copy the online representation network into the object discovery network (e.g.\ $\lambda_\tau = 1$ every 100 epochs and $\lambda_\tau = 0$ otherwise).
The advantage of the second scheme is computational: if the object discovery network does not change between updates, the segments for training the object representation networks can be cached, removing the need to evaluate the object discovery network at every iteration.

\vspace{1em} \noindent \textbf{Pretraining details.}
We train object discovery and representation networks (Odin) on ImageNet \cite{russakovsky2015imagenet} for 1000 epochs, using a ResNet-50 \cite{he2016deep} or Swin Transformer \cite{liu2021swin} backbone equipped with Feature Pyramid Networks (FPN; \cite{lin2017feature}) as the feature extractor $f$. The FPN takes as input the hierarchy of latent vectors output by the ResNet or Swin backbone, and progressively upsamples them while adding information from intermediate feature arrays, yielding high-level and high-resolution representations of the input image. We use the highest-resolution output of the FPN (subsampling the image by a factor of 4) as the array of hidden vectors $\vh$.

After pretraining, we discard the target and object discovery networks, and use only the online object representation network for evaluation, facilitating the comparison to other methods which have their own means of learning the model. 

\subsection{Evaluating object discovery and representation} Having trained object representation networks using the Odin framework, we evaluate the quality of their representation by fine-tuning them for object detection and instance segmentation on COCO, and segmentatic segmentation  on PASCAL and Cityscapes. For consistency with prior work we retain only the pretrained backbone (ResNet or Swin transformer) for transfer learning, discarding feature pyramid networks and projection heads.

\vspace{1em} \noindent \textbf{Object detection and instance segmentation.}
For instance segmentation we use Mask-RCNN \cite{he2017mask} while for object detection we report results for Mask-RCNN and \fcos.
Both methods are equipped with feature pyramid networks \cite{lin2017feature} and cross-replica batch-norm \cite{peng2018megdet}.
For Mask-RCNN we adopt the Cloud TPU implementation \cite{cloud_tpu_mask_rcnn} and use it without modification.
\fcos is our implementation of a single-stage detector based on FCOS \cite{tian2019fcos}, and improved with IoU prediction \cite{wu2020iou}, ATSS \cite{zhang2020bridging} and T-Head \cite{feng2021tood};
full details are available in the appendix.  
We follow the common transfer setup and evaluate on COCO \cite{lin2014microsoft} --
the pretrained network is used to initialize the backbone of a Mask-RCNN or \fcos model,
which is then fine-tuned on the \texttt{train2017} set, and report bounding-box AP (\apbbox{}) and mask AP (\apmask{}) on the \texttt{val2017} set. We use two standard training schedules: 12 epochs and 24 epochs \cite{he2019momentum}.

\vspace{1em} \noindent \textbf{Semantic segmentation with FCN.} 
Following \cite{he2019momentum} we initialize the backbone of a fully-convolutional network (FCN, \cite{long2015fully}) with our model. For PASCAL \cite{everingham2015pascal}, we fine-tune on the \texttt{train\_aug2012} set for 45 epochs and report the mean intersection over union (mIoU) on the \texttt{val2012} set. For Cityscapes \cite{cordts2016cityscapes}, we fine-tune on the \texttt{train\_fine} set for 160 epochs and evaluate on the  \texttt{val\_fine} set.

\vspace{1em} \noindent \textbf{Object discovery on COCO.} We wish to assess whether our representations uncover the structure of real-world scenes during self-supervised pretraining. Simply visualizing saliency maps induced by the model only weakly tests this ability however \cite{dwibedi2021little}, hence we use the COCO dataset comprised of complex scenes and human-annotated object segments. Specifically, we evaluate models on COCO images, cluster their features, and measure the overlap between the resulting segments and human-annotated ones. Given the diversity of object scales in COCO, we run multiple $K$-means segmentations (for $K$ in [1, 2, $\dots$, 128]) on the same set of latents, resulting in 255 object proposals which we resize to the input image resolution.

For each ground-truth segment $\vg_t$ we compute the overlap with all proposals $\vm_k$ using their intersection-over-union (IoU), and record the ``best overlap'' by taking the maximum across proposals. Averaging this metric across ground-truth segments, we obtain the ``average best overlap'' (ABO) metric \cite{arbelaez2014multiscale}, and computing the fraction of ``best overlaps'' greater than 50\% yields the ``object recovery'' metric \cite{cho2015unsupervised}. We then average each of these metrics across images. 

\vspace{1em} \noindent \textbf{Video object segmentation on DAVIS.} As a further test of scene understanding, we assess whether learned representations can continue to recognize parts of an object as they evolve over time. Video object segmentation, specifically in its semi-supervised setting, captures this ability, which we evaluate on the DAVIS'17 benchmark \cite{perazzi2016benchmark}. Having evaluated a learned representation on a video independently across frames, we segment these features with nearest neighbor matching from frame to frame, given a segmentation of the first frame. In this way, the segmentation is propagated according to the similarity of the representation across space and time.

\section{Experiments}

\subsection{Transfer learning}

Our first goal is to assess whether strong transfer learning performance can be obtained without resorting to prior knowledge of scene segmentations. To that end we train a ResNet-50 on ImageNet for 1000 epochs using the proposed Odin framework, and transfer it to object detection and instance segmentation on COCO, and semantic segmentation on PASCAL and Cityscapes.

\begin{table}[t]
\caption{\textbf{Transfer to COCO object detection and instance segmentation with Mask-RCNN:} all methods pretrain a ResNet-50 on ImageNet before fine-tuning on COCO with Mask-RCNN for 12 epochs (1\x \ schedule) or 24 epochs (2\x \ schedule). We report average precision on object detection (\apbbox{~}) and instance segmentation (\apmask{~})}
\small
\begin{tabularx}{\linewidth}{c *{6}{Y}}
 \multicolumn{2}{c}{pretraining}  
 & \multicolumn{2}{c}{fine-tune 1\x}  
 & \multicolumn{2}{c}{fine-tune 2\x}
 \\
\cmidrule(lr){1-2} \cmidrule(lr){3-4} \cmidrule(l){5-6} 
Method & Knows obj? & \apbbox{~} & \apmask{~} & \apbbox{~} & \apmask{~} \\
\midrule
Supervised & \doesntknowobj & 39.6 & 35.6 & 41.6 & 37.6 \\
VADeR \cite{pinheiro2020unsupervised} & \doesntknowobj  & 39.2 & 35.6 & - & - \\
MoCo \cite{he2019momentum} & \doesntknowobj  & 39.4 & 35.6 & 41.7 & 37.5  \\
SimCLR \cite{chen2020simple} &  \doesntknowobj & 39.7 & 35.8 & 41.6 & 37.4 \\ 
MoCo v2 \cite{chen2020improved} &  \doesntknowobj & 40.1 & 36.3 & 41.7 & 37.6 \\ 
InfoMin \cite{tian2020makes} & \doesntknowobj  & 40.6 & 36.7 & 42.5 & 38.4 \\ 
DeepCluster-v2 \cite{caron2020unsupervised} & \doesntknowobj  & 41.1 & 37.1 & - & - \\ 
DINO \cite{caron2021emerging} & \doesntknowobj & 41.2 & 37.1 & 42.3 & 38.1 \\  
PixPro \cite{xie2020propagate} & \doesntknowobj  & 41.4 & - &  - &  - \\ 
BYOL \cite{grill2020bootstrap} & \doesntknowobj  & 41.6 & 37.2 & 42.4 & 38.0 \\ 
SwAV \cite{caron2020unsupervised} & \doesntknowobj  & 41.6 & 37.8 & - & - \\ 
 ReLIC v2 \cite{tomasev2022pushing} & \knowsobj & 42.5 & 38.0 & 43.3 & 38.6 \\ 
 DetCon$_{B}$ \cite{henaff2021efficient} & \knowsobj & 42.7 & 38.2 & 43.4 & 38.7 \\
\midrule
\textbf{Odin}  & \doesntknowobj & \textbf{42.9} & \textbf{38.4} & \textbf{43.8} & \textbf{39.1}  \\ 
\end{tabularx}
\label{tab:prior_art_r50}
\vspace{-1.em}
\end{table}

\vspace{1em} \noindent \textbf{Object detection and instance segmentation on COCO.} Self-supervised learning has made steady gains on transfer learning from ImageNet to COCO, with a majority of methods surpassing supervised pretraining. The top-performing methods are ReLIC v2 and DetCon$_B$ which make heavy use of saliency or segmentation information in their learning paradigm. DetCon uses the same learning objective as Odin, but relies on a hand-crafted image segmentation algorithm \cite{felzenszwalb2004efficient} applied to the pixel lattice rather than a learned object discovery network. ReLIC v2 does not use segmentation information explicitly in its objective, but uses a hand-crafted saliency network to separate objects from their background in the data-augmentation pipeline. Both represent a step-change in performance relative to previous methods. Odin, which instead derives segmentations from its own learned representations, surpasses both of these methods (Table \ref{tab:prior_art_r50}). 

A recent self-supervised method, DINO \cite{caron2021emerging}, reports high-quality unsupervised segmentations, however it appears to do so at the cost of object representation. We fine-tune the publicly available ResNet checkpoint in our framework, and find it underperforms relative to simple methods such as BYOL. Other SSL methods such as SwAV and DeepCluster-v2 \cite{caron2020unsupervised} which cluster representations \textit{across} images rather than within also underperform in this setting. 

\begin{table}[t]
\caption{\textbf{Transfer to PASCAL and Cityscapes semantic segmentation with fully convolutional networks:} all methods pretrain a ResNet-50 on ImageNet before fine-tuning for semantic segmentation on PASCAL or Cityscapes, and report the mean intersection-over-union}
\small
\begin{tabularx}{\linewidth}{c *{4}{Y}}
Method & Knows obj? & PASCAL & Cityscapes  \\
\midrule
Supervised & \doesntknowobj &  74.4 & 74.9 \\
BYOL \cite{grill2020bootstrap} & \doesntknowobj   & 75.7 & 74.6\\ 

DeepCluster-v2 \cite{caron2020unsupervised} & \doesntknowobj & 75.8 & 76.8 \\ 
SwAV \cite{caron2020unsupervised} & \doesntknowobj   & 76.0 & 76.2  \\ 

DINO \cite{caron2021emerging} & \doesntknowobj   & 76.9 & 75.6 \\ 
DetCon$_{B}$ \cite{henaff2021efficient} & \knowsobj & 77.3 & 77.0 \\ 
ReLIC v2 \cite{tomasev2022pushing} & \knowsobj & 77.9 & 75.2 \\ 
\midrule
\textbf{Odin}  & \doesntknowobj & \textbf{78.6} & \textbf{77.1}   \\ 
\end{tabularx}
\label{tab:prior_art_seg}
\vspace{-1.em}
\end{table}

\vspace{1em} \noindent \textbf{Semantic segmentation on PASCAL and Cityscapes.}
We assess the generality of these results by transferring them to two separate datasets and tasks, semantic segmetation on PASCAL and Cityscapes. Similarly to when transferring to COCO, DetCon and ReLIC v2 substantially outperform supervised and BYOL pretraining, confirming the utility of prior knowledge about segmentation and saliency. In this case as well, Odin successfully recovers this knowledge and surpasses both methods in a fully learned manner  (Table \ref{tab:prior_art_seg}).

In this setting DINO performs better, surpassing BYOL, possibly because semantic segmentation on PASCAL, which contains only 20 classes compared with 80 in COCO, weights object discovery more than object representation---isolating objects from the background rather than distinguishing object classes from each other. 
Nevertheless, Odin surpasses it as well, indicating that it achieves a better trade-off between object representation and discovery.

\vspace{1em} \noindent \textbf{Transfer learning with high-performance architectures.} While Mask-RCNN has become a standard method for evaluating the quality of object-level representations, we asked whether the performance gains afforded by the Odin framework persisted with more sophisticated models. For this we turned to our \fcos implementation, whose supervised baseline surpasses Mask-RCNN by 4.6\% \apbbox{}. In this setting as well, Odin surpasses the supervised baseline and DINO (+1.3\% \apbbox{}, Table \ref{tab:prior_art_fcos}, \nth{1} column).  


\begin{table}[t]
\vspace{1.em}
\caption{\textbf{Transfer to COCO object detection with \fcos:} all methods pretrain on ImageNet before fine-tuning on COCO with \fcos for 30 epochs, and report average precision on object detection (\apbbox{~}).}
\small
\begin{tabularx}{\linewidth}{c *{5}{Y}}
Pretraining & Knows obj? & ResNet-50 & Swin-T & Swin-S  \\
\midrule
Supervised & \doesntknowobj & 44.2 & 46.7 & 48.3 \\
DINO \cite{caron2021emerging}  & \doesntknowobj  & 44.3 & - & - \\ 
MOBY \cite{xie2021self}  & \doesntknowobj  & - & 47.6 & - \\ 
DetCon$_{B}$ \cite{henaff2021efficient} & \knowsobj & 45.4 & 48.4 & \textbf{50.4} \\ 
\midrule
\textbf{Odin}  & \doesntknowobj & \textbf{45.6} & \textbf{48.5} & \textbf{50.4} \\ 
\end{tabularx}
\label{tab:prior_art_fcos}
\vspace{-1.em}
\end{table}

Swin transformers appear as a compelling candidate for general-purpose vision architectures, surpassing ResNet's in a variety of tasks \cite{liu2021swin}. Despite the almost universal success of self-supervised pretraining in improving the transfer learning performance of ResNet architectures, similar results have yet to become widespread for Swin transformers. 

We therefore pretrain Swin-T and Swin-S transformers on ImageNet using Odin, and transfer them to COCO object detection using \fcos. We also evaluate a pre-trained Moby checkpoint in the same setting. Moby pretraining marginally improves the performance of a Swin-T, whereas Odin furthers these gains (+1.8\% \apbbox{}, Table \ref{tab:prior_art_fcos}, \nth{2} column). The benefits of Odin pretraining are emphasized when pretraining and transferring the larger Swin-S backbone (+2.1\% \apbbox{}, Table \ref{tab:prior_art_fcos}, \nth{3} column). 

We return to our original question of whether Odin has successfully recovered knowledge of scene structure by pretraining ResNet-50, Swin-T, and Swin-S with DetCon (which uses a hand-crafted segmentation algorithm instead of the object discovery network \cite{felzenszwalb2004efficient,henaff2021efficient}), and transferring them with \fcos. We find Odin to match or slightly surpass their performance, confirming our previous results.

\subsection{Object discovery in COCO}
\label{sec:res-obj-discovery}

\begin{figure}[h!]
\centering
\includegraphics[width=\textwidth,height=0.89\textheight]{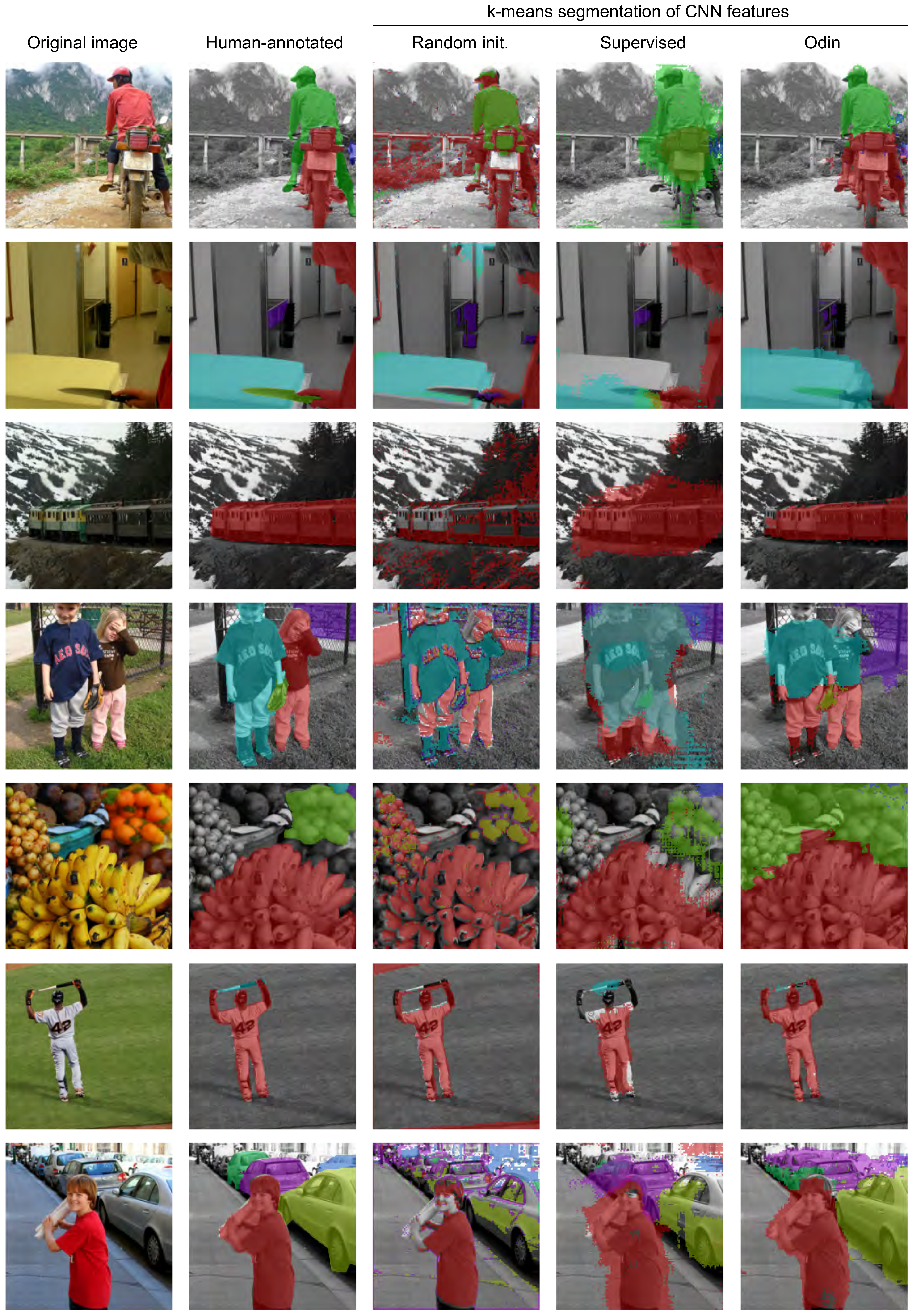}
\vspace{-2em}
\caption{Object discovery with Odin. \nth{1} column: original image, \nth{2}: human-annotated COCO segmentations, \nth{3}, \nth{4}, \nth{5}: segmentations obtained from k-means clustering on randomly intialized, ImageNet-supervised, and Odin-trained features, respectively.}
\label{fig:segmentations}
\end{figure}

We have found thus far that Odin surpasses the transfer learning performance of state-of-the-art self-supervised methods which rely on prior knowledge of scene segmentations, suggesting it has derived this knowledge from the data itself. In this section, we directly evaluate the extent to which Odin has discovered objects in real-world scenes. We extract Odin features from COCO images, cluster them, and visualize the resulting segments (Figure \ref{fig:segmentations}). Comparing unsupervised object proposals (last column) to human-annotated segments (\nth{2} column) we see that Odin recovers a reasonable decomposition of real-world scenes: figures are separated from their background, small objects are isolated, and even different instances of the same class (such as cars, last row) are roughly separated from one-another. Failure modes, such as grouping multiple fruit---or two shirts---together, reflect the ambiguity of unsupervised object discovery.

Comparing these proposals to those obtained from a randomly-initialized network (\nth{3} column) and an ImageNet supervised one (\nth{4} column), we appreciate the benefits of learning with the Odin framework. Both of these networks make erroneous proposals, failing to delineate object boundaries, or lacking the coherence and locality of real-world objects. We quantify this difference by evaluating the average best overlap (ABO) and fraction of recovered objects (OR) of the segments derived from each network. Consistently with the qualitative results, Odin strongly surpasses both baselines in all metrics (Table \ref{tab:object_discovery}, left). 

We also evaluate the accuracy of a recently-proposed self-supervised method, DINO, which specializes in object discovery. In this challenging task of discovering multiple objects in an unsupervised setting, we find that it underperforms relative to Odin. We test Odin in two regimes, one using the ResNet and FPN used for pretraining, the other with the ResNet only. Although its performance degrades slightly with the lower-resolution ResNet, it continues to outperform all other methods in all metrics. In particular, Odin surpasses DetCon by a large margin (+7\% ABO, +16\% OR), indicating that it has discovered more relevant image structure than the hand-crafted segmentations used in DetCon.

Finally, we note that the DINO method was primarily designed for use with vision transformers \cite{dosovitskiy2020image}. We therefore train a ViT-B/8 (as in DINO) on ImageNet for 100 epochs using the Odin framework (all other parameters unchanged). In this setting we find Odin to achieve compelling results, surpassing the high-resolution ResNet-FPN, and a supervised and DINO-pretrained vision-transformer (Table \ref{tab:object_discovery}, right). Figure 
A.1. illustrates that Odin seems particularly effective at discovering small objects and differentiating instances of the same class. In sum, Odin provides a powerful means of discovering objects in real-world scenes irrespective of the architecture used. 

\begin{table}[t]
\caption{\textbf{Object discovery on COCO:} all methods pretrain on ImageNet before evaluating object discovery on COCO in an unsupervised manner, reporting average best overlap of instance masks (ABO$^i$) and categorical masks (ABO$^c$), and average object recovery (OR). By default we retain only the pretrained ResNet-50 from Odin's feature extractor, such that all methods are matched in their architecture. \textbf{Odin}$^\dagger$ denotes the model equipped with the FPN used during training. ResNet's use 1024\x 1024 images for evaluation with a stride of 32, yielding a 32\x32 feature grid. ViT's use 448\x448 resolution with a patch size of 8, yielding 56\x56 feature grids.} 
\vspace{-.5em}
\small
\begin{tabularx}{\linewidth}{c *{7}{Y}}
& \multicolumn{3}{c}{ResNet-50} & \multicolumn{3}{c}{ViT-B}  \\
\cmidrule(lr){2-4}  \cmidrule(lr){5-7} 
Pretraining & ABO$^i$ & ABO$^c$ & OR & ABO$^i$ & ABO$^c$ & OR  \\
\midrule
Random init & 28.1 & 33.6 & 17.0 & 27.8 & 33.6 & 17.5 \\
DetCon$_{B}$ \cite{henaff2021efficient} & 34.1 & 40.0 & 20.4 & - & - & - \\

Supervised & 35.8 & 41.1 & 23.8 & 43.9 & 53.6 & 41.9 \\

DINO \cite{caron2021emerging} & 38.3 & 46.5 & 30.8 & 42.7 & 51.7 & 39.7 \\


\midrule
\textbf{Odin} & \textbf{41.5} & \textbf{48.6} & \textbf{36.5} & \textbf{45.9} & \textbf{53.9} & \textbf{44.1}\\
\hspace{0.125em} \textbf{Odin}$^\dagger$ & \textbf{43.0} & \textbf{53.0} & \textbf{42.3} \\ 
\end{tabularx}
\label{tab:object_discovery}
\end{table}

\subsection{Video Object Segmentation}
We evaluate on the DAVIS'17 benchmark \cite{perazzi2016benchmark} following the experimental setup and nonparametric inference method of DINO \cite{caron2021emerging}.
Given a video and a segmentation of its first frame, we propagate the segmentation between consecutive frames by nearest neighbor matching of the extracted representation.
In Table \ref{tab:video_seg} we evaluate random, supervised, and our self-supervised representations with the ResNet-50 architecture.
This evaluation does not fine-tune or train on the DAVIS benchmark, and so accuracy is a measure of object representation, as the fixed representation must support segmentation of the novel objects in these held-out videos. Consistently with our previous results, Odin strongly surpasses supervised pre-training in all metrics. 

\begin{table}[t]
\caption{
    \textbf{Video Object Segmentation on DAVIS'17:} 
    we evaluate representation quality for video segmentation by nearest neighbor inference. 
    We report the standard region $\mathcal{J}$ and contour $\mathcal{F}$ metrics and their mean.
    All representations are trained on ImageNet then evaluated without fine-tuning on the 30 validation videos of DAVIS'17.
    $\text{Odin}^{\dagger}$ includes a feature pyramid network to reduce the output stride from $32\times$ to $8\times$.
}
\vspace{-.5em}
\small
\centering
\begin{tabularx}{0.7\linewidth}{c *{4}{Y}}
Pretraining & $(\mathcal{J} \& \mathcal{F})_m$ & $\mathcal{J}$ & $\mathcal{F}$ \\
\midrule
Random                     & 15.2 & 15.9 & 14.6 \\  
Supervised                 & 27.0 & 33.0 & 20.9 \\  
\midrule        
Odin                       & \textbf{35.6} & \textbf{41.3} & \textbf{29.9} \\ 
\textbf{Odin$^{\dagger}$}  & \textbf{54.1} & \textbf{54.3} & \textbf{53.9} \\ 
\end{tabularx}
\label{tab:video_seg}
\vspace{-1.5em}
\end{table}

\subsection{Ablations and analysis}

What components are necessary for driving Odin's ability to represent and discover objects? We systematically vary the two hyper-parameters governing the behavior of the object discovery network: the number of segments used for learning, and the schedule used for updating the network (Table \ref{tab:ablation}). Starting with the number of segments $K$ we find object discovery degrades substantially when using too coarse segmentations (e.g.\ $K = 8$). However, given a fine enough segmentations ($K$ greater than 16) its performance is stable. 

Regarding the rate at which the object discovery network is updated, we find both schemes to be viable: continuously updating the network leads to slightly better representations, whereas discrete updates lead to slightly better object discovery. The advantage of the later scheme is that the computational cost of the object discovery network becomes negligible, as it only needs to be evaluated every 100 epochs and resulting segmentations cached in-between.



\begin{table}[t]
\vspace{1.em}
\caption{\textbf{Ablating the components of Odin:} We use the variant of Odin equipped with FPN for object discovery. Transfer learning is performed with the ResNet backbone only. $K$ denotes the number of segments obtained through $K$-means during pretraining}
\vspace{-.5em}
\small
\begin{tabularx}{\linewidth}{c *{6}{Y}}
\multicolumn{2}{c}{Odin pretraining} & \multicolumn{2}{c}{Object discovery} & \multicolumn{2}{c}{Mask-RCNN transfer}  \\
\cmidrule(lr){1-2} \cmidrule(lr){3-4}  \cmidrule(lr){5-6} 
$\hspace{1.75em}K\hspace{1.25em}$ & update sched. & ABO$^i$  & OR & \apbbox{~} & \apmask{~} \\
\midrule

\hspace{0.5em}8 & every 100 ep.           & 38.3 & 34.6 & 42.6 & 38.1 \\ 
16              & every 100 ep.           & 43.0 & \textbf{42.3} & 42.6 & 38.1 \\ 
32              & every 100 ep.           & \textbf{43.1} & 42.0 & 42.5 & 38.0 \\ 
16              & $\lambda_\tau=10^{-2}$  & 41.0 & 39.5 & 42.5 & 38.1 \\ 
16              & $\lambda_\tau=10^{-3}$  & 41.3 & 39.9 & \textbf{42.9} & \textbf{38.4} \\ 
16              & $\lambda_\tau=10^{-4}$  & 41.6 & 40.1 & 42.6 & 38.1 \\ 

\end{tabularx}
\label{tab:ablation}
\vspace{-2.em}
\end{table}


\section{Conclusions}

We have presented Odin, a new approach to self-supervised training which couples object discovery and representation learning. The resulting framework benefits from the same representation quality as methods which utilize explicit priors about image segmentation \cite{henaff2021efficient,tomasev2022pushing}, while deriving this knowledge from the data itself. The result is a simpler and more generally-applicable learning paradigm, and leads to state-of-the-art performance on a range of transfer learning tasks. 


In this work, we have shown the utility of coupling representation learning and object discovery, for transfer learning and unsupervised scene understanding. Nevertheless, we have presented a single instance of this coupling, and there remain several open questions around how best to tie them together. This may require greater integration of the learning procedure and architecture---for example our self-supervised algorithm learns mask-pooled features which are different from those used in downstream tasks. The learning dynamics of Odin also warrant further investigation, as well as the objective used for representation learning. Recent work has revived interest in masked-autoencoding \cite{dosovitskiy2020image,bao2021beit,he2021masked} and masked-distillation \cite{baevski2022data2vec} as viable alternatives to contrastive learning. Odin, by proposing to leverage learned representations in the design of iteratively refined self-supervised tasks, is well positioned to benefit them as well. 

\clearpage
%

\bibliographystyle{splncs04}
\bibliography{bib}


\newpage

\appendix
\counterwithin{table}{section}
\counterwithin{figure}{section}

\section{Appendix}

\subsection{Implementation: data pre-processing}
\label{sec:app-data}

\noindent \textbf{Self-supervised pretraining.} Each image is randomly augmented twice, resulting in two views $\vv^1$ and $\vv^2$. The augmentations are constructed as compositions of the following operations, each applied with a given probability:
\begin{enumerate}
\item random cropping: a random patch of the image is selected, whose area is uniformly sampled in $[0.08 \cdot \mathcal{A}, \mathcal{A}]$, where $\mathcal{A}$ is the area of the original image, and whose aspect ratio is logarithmically sampled in $[3/4, 4/3]$. The patch is then resized to 224 \x 224 pixels using bicubic interpolation;
\item horizontal flipping;
\item color jittering: the brightness, contrast, saturation and hue are shifted by a uniformly distributed offset;
\item color dropping: the RGB image is replaced by its grey-scale values;
\item gaussian blurring with a 23\x23 square kernel and a standard deviation uniformly sampled from $[0.1, 2.0]$;
\item solarization: a point-wise color transformation $x \mapsto x \cdot \mathds{1}_{x < 0.5} + (1 - x) \cdot \mathds{1}_{x \ge 0.5}$ with pixels $x$ in $[0, 1]$.
\end{enumerate}
\noindent The augmented images $\vv^1$ and $\vv^2$ result from augmentations sampled from distributions $\mathcal{T}_1$ and~$\mathcal{T}_2$ respectively. These distributions apply the primitives described above with different probabilities, and different magnitudes. The following table specifies these parameters for the BYOL framework \cite{grill2020bootstrap}, which we adopt without modification. 
\begin{table}[ht]
    \small
    \centering
    \begin{tabular}{l c c }
    Parameter                           & $\hspace{1em} \mathcal{T}_1 \hspace{1em}$ & $\hspace{1em} \mathcal{T}_2 \hspace{1em}$ \\ \hline
    Random crop probability             & \multicolumn{2}{c}{1.0} \\
    Flip probability                    & \multicolumn{2}{c}{0.5} \\
    Color jittering probability         & \multicolumn{2}{c}{0.8} \\
    Color dropping probability          & \multicolumn{2}{c}{0.2} \\
    Brightness adjustment max           & \multicolumn{2}{c}{0.4} \\
    Contrast adjustment max             & \multicolumn{2}{c}{0.4} \\
    Saturation adjustment max           & \multicolumn{2}{c}{0.2} \\
    Hue adjustment max                  & \multicolumn{2}{c}{0.1} \\
    Gaussian blurring probability       & $1.0$ & $0.1$ \\
    Solarization probability            & $0.0$ & $0.2$ \\
    \end{tabular}
\end{table}

\noindent The ``spanning view'' $\vv^0$ is chosen as the smallest crop spanning the spatial extent of $\vv^1$ and $\vv^2$. When training ResNet or Swin, we resize $\vv^0$ to 448\x448 resolution. When training vision transformers, we resize it to 224\x224 resolution.

\vspace{0.5em} \noindent \textbf{Transfer to COCO.} Resolutions used for Mask-RCNN and \fcos are 1024\x1024 and 800\x1024, respectively. During testing, an image is resized by a factor $s$ while preserving the aspect ratio, such that it is tightly contained inside the target resolution, and then padded. When fine-tuning, the image is rescaled by a factor of $u \cdot s$ where $u$ is uniformly sampled in $[0.8, 1.25]$, and is then cropped or padded to the target resolution.

\vspace{0.5em} \noindent \textbf{Transfer to PASCAL.} During training, images are randomly flipped and scaled by a factor in $[0.5, 2.0]$. Training and testing are performed with 513\x513-resolution images.  

\vspace{0.5em} \noindent \textbf{Transfer to Cityscapes.} During training, images are randomly flipped horizontally and scaled by a factor in $[0.5, 2.0]$. Training is performed on 769\x769-resolution images and testing on 1025\x2049-resolution images. 

\subsection{Implementation: details of \fcos}
\label{sec:app-fcos}

Here we describe \fcos, a fully convolutional single-stage object detector based on FCOS \cite{tian2019fcos} and its improvements \cite{wu2020iou,zhang2020bridging,feng2021tood}.

At inference time, as in FCOS \cite{tian2019fcos}, an image is ingested by a backbone network such as ResNet-50 or Swin transformer, followed by a feature pyramid network \cite{lin2017feature}
which produces dense feature maps at various scales (feature pyramid levels).
Each feature map is processed independently by prediction heads with shared weights,
producing three sets of dense outputs for each pyramid level:
classification logits ($cls$), quality logits ($qual$) and bounding box parameters.
Each location in each output map corresponds to a detection
with the regressed bounding box parameters
(parametrized as distances to the 4 edges divided by the level's stride),
where the score for the detection of class $c$ is computed as
$\sqrt{\sigma(cls_c) \times \sigma(qual)}$,
and $\sigma$ is the sigmoid function.
Non-maximum suppression is performed to produce the final set of detections.

The network is trained to
1) predict the correct class by minimizing the focal loss \cite{lin2017focal} as in FCOS \cite{tian2019fcos},
2) regress the bounding box parameters for positive samples by minimizing the GIoU loss \cite{rezatofighi2018generalized} as in FCOS \cite{tian2019fcos}, and
3) predict correct detections using the quality logits as in \cite{wu2020iou} by minimizing the binary cross-entropy loss.
The positive samples are defined through the assignment of the dense predictions
to ground truth boxes via ATSS \cite{zhang2020bridging}.

For all components and parameters of the method we use the default settings from
the respective papers.
Slight departures consist of using
1) a lower resolution ($800 \times 1024$ instead of more standard $800 \times 1333$),
2) the cross-replica batch-norm \cite{peng2018megdet} in backbones where applicable, and
3) the T-Head for all three prediction heads while \cite{feng2021tood} does not use it for the quality branch.

\subsection{Implementation: optimization}
\label{sec:app-optimization}

\vspace{0.5em} \noindent \textbf{Self-supervised pretraining.} We pretrain ResNet-50 and Swin-Transformers on ImageNet for 1000 epochs using the LARS optimizer \cite{you2017large} with a batch size of 4096 split across 128 Cloud TPU v3 workers. We adopt the optimization details of BYOL, scaling the learning rate linearly with the batch size and decaying it according to a cosine schedule. The base learning rate is 0.2 and the weight decay is $1.5 \cdot 10^{-6}$. When pretraining Swin transformers we additionally use gradient clipping with a maximum norm of 1. The contrastive loss temperature $\alpha$ is 0.1. 

We pretrain vision transformers (ViT) on ImageNet for 300 epochs using the Adam optimizer \cite{kingma2014adam} with a batch size of 2048 split across 256 Cloud TPU v3 workers. We use a learning rate of $3 \cdot 10^{-4}$, a weight decay of 0.1, and momentum parameters $\beta_1 = 0.9$ and $\beta_2 = 0.95$.

When training ResNet-50 and Swin-Transformers, we fuse intermediate feature maps into a single high-resolution latent array using Feature Pyramid Networks, and segment these features using their projections $\vz^0$. When training vision transformers, we use the hidden vectors $\vh^0$ for segmentation. 

\vspace{0.5em} \noindent \textbf{Transfer to COCO with Mask-RCNN.} We fine-tune with stochastic gradient descent, increasing the learning rate linearly for the first 500 iterations and dropping twice by a factor of 10, after $\frac{2}{3}$ and $\frac{8}{9}$ of the total training time, following \cite{wu2019detectron2}. We use a base learning rate of 0.3, a momentum of 0.9, a weight decay of 4$\cdot$10$^{-5}$, and a batch size of 64 images split across 16 workers. 

\vspace{0.5em} \noindent \textbf{Transfer to COCO with \fcos.}
The network is trained for 30 epochs with batch size 128 split across 16 workers,
with AdamW \cite{loshchilov2019decoupled}, weight decay $10^{-4}$ and
base learning rate of $10^{-3}$.
The learning rate rises linearly for $\frac{1}{4}$ of an epoch,
and is dropped twice by a factor of 10,
after $\frac{2}{3}$ and $\frac{8}{9}$ of the total training time.

\vspace{0.5em} \noindent \textbf{Transfer to PASCAL.} We fine-tune for 45 epochs with stochastic gradient descent, with a batch size of 16 and weight decay of $10^{-4}$. The learning rate is 0.02 and dropped by a factor of 10 at the \nth{70} and \nth{90} percentiles.

\vspace{0.5em} \noindent \textbf{Transfer to Cityscapes.} We fine-tune for 160 epochs with stochastic gradient descent and a Nesterov momentum of 0.9, using a batch size of 2 and weight decay of $10^{-4}$. The initial learning rate is 0.005 and dropped by a factor of 10 at the \nth{70} and \nth{90} percentiles.

\subsection{Implementation: evaluating object discovery}
\label{sec:app-object-discovery}

We extract the central crop of COCO images and resize them to a target resolution of 1024\x1024 for ResNet models, and 448\x448 for vision transformers. These images are encoded by the feature extractor, resulting in a 32\x32 feature grid for ResNet, 256\x256 for ResNet equipped with FPN, and 56\x56 for the vision transformer. For ease of comparison with other forms of pretraining, when evaluating ViT and ResNet we use the final hidden layer $\vh$ (with 768 and 2048 channels, respectively) for unsupervised segmentation. When using the ResNet equipped with FPN, we use the projections $\vz$. In all cases, vectors are $L^2$ normalized before applying k-means clustering. 

For each ground-truth segment $\vg_t$ we compute the overlap with all proposals $\vm_k$ using their intersection-over-union (IoU), and record the best overlap (BO) by taking the maximum across proposals:
\begin{align}
    \textrm{IoU}( \vg_t, \vm_k ) & = \frac{ \sum_{i,j} \min( \vg_t, \vm_k )[i, j] }{ \sum_{i,j} \max( \vg_t, \vm_k )[i, j] } \\ 
    \textrm{BO}( \vg_t ) & = \max_k \textrm{IoU}( \vg_t, \vm_k ).
\end{align}
For a given image with $T$ ground-truth segments, we obtain the ``average best overlap'' (ABO) metric \cite{arbelaez2014multiscale} by averaging BO across ground-truth segments, and the ``object recovery'' metric \cite{cho2015unsupervised} by computing the fraction of ``best overlaps'' greater than 50\%:
\begin{align}
    \textrm{ABO} & = \frac{1}{T} \sum^T_{t=1} \textrm{BO}( \vg_t ) \\ 
    \textrm{OR}  & = \frac{1}{T} \sum^T_{t=1} \mathds{1}_{\textrm{BO}( \vg_t ) > 0.5} 
\end{align}
The ABO$^i$ and OR metrics use instance-based masks as ground-truth segments $\vg_t$. The ABO$^c$ metric merges instances of the same class into the same ground-truth segment, before applying the same analysis. We average each of these metrics across images. 

In Table \ref{tab:object_discovery} we perform this analysis using the original COCO masks, without modification. We notice however that some COCO masks are very small, making them barely noticeable perceptually. We verified that these masks do not bias our results by repeating the analysis while excluding these masks. Specifically, we limit the number of masks to 16 per image, and remove masks whose area is smaller than 100 pixels (in a 224\x224-resolution image). Table \ref{tab:object_discovery_raw} shows that, after this preprocessing of COCO masks, our results are very similar, and our conclusions are unchanged.

\begin{table}[h]
\caption{\textbf{Object discovery on COCO, with ground-truth mask filtering:} we apply the same object discovery analysis as in Table 4, after filtering COCO masks to remove very small segments (see section \ref{sec:app-object-discovery})} 
\vspace{-.5em}
\small
\begin{tabularx}{\linewidth}{c *{7}{Y}}
& \multicolumn{3}{c}{ResNet-50} & \multicolumn{3}{c}{ViT-B}  \\
\cmidrule(lr){2-4}  \cmidrule(lr){5-7} 
Pretraining & ABO$^i$ & ABO$^c$ & OR & ABO$^i$ & ABO$^c$ & OR  \\
\midrule
Random init & 29.6 & 33.6 & 16.7 & 29.5 & 34.2 & 18.0 \\ 
Supervised & 39.5 & 42.4 & 25.6 & 42.9 & 49.7 & 36.8 \\ 
DINO & 40.5 & 46.0 & 30.3 & 42.7 & 48.3 & 35.7 \\ 
\midrule
\textbf{Odin} & \textbf{44.7} & \textbf{49.0} & \textbf{38.4} & \textbf{49.7} & \textbf{54.7} & \textbf{48.2} \\ 
\hspace{0.125em} \textbf{Odin}$^\dagger$ & \textbf{47.2} & \textbf{53.9} & \textbf{46.2} \\ 
\end{tabularx}
\label{tab:object_discovery_raw}
\end{table}

\begin{figure}[h!]
\centering
\includegraphics[width=\textwidth]{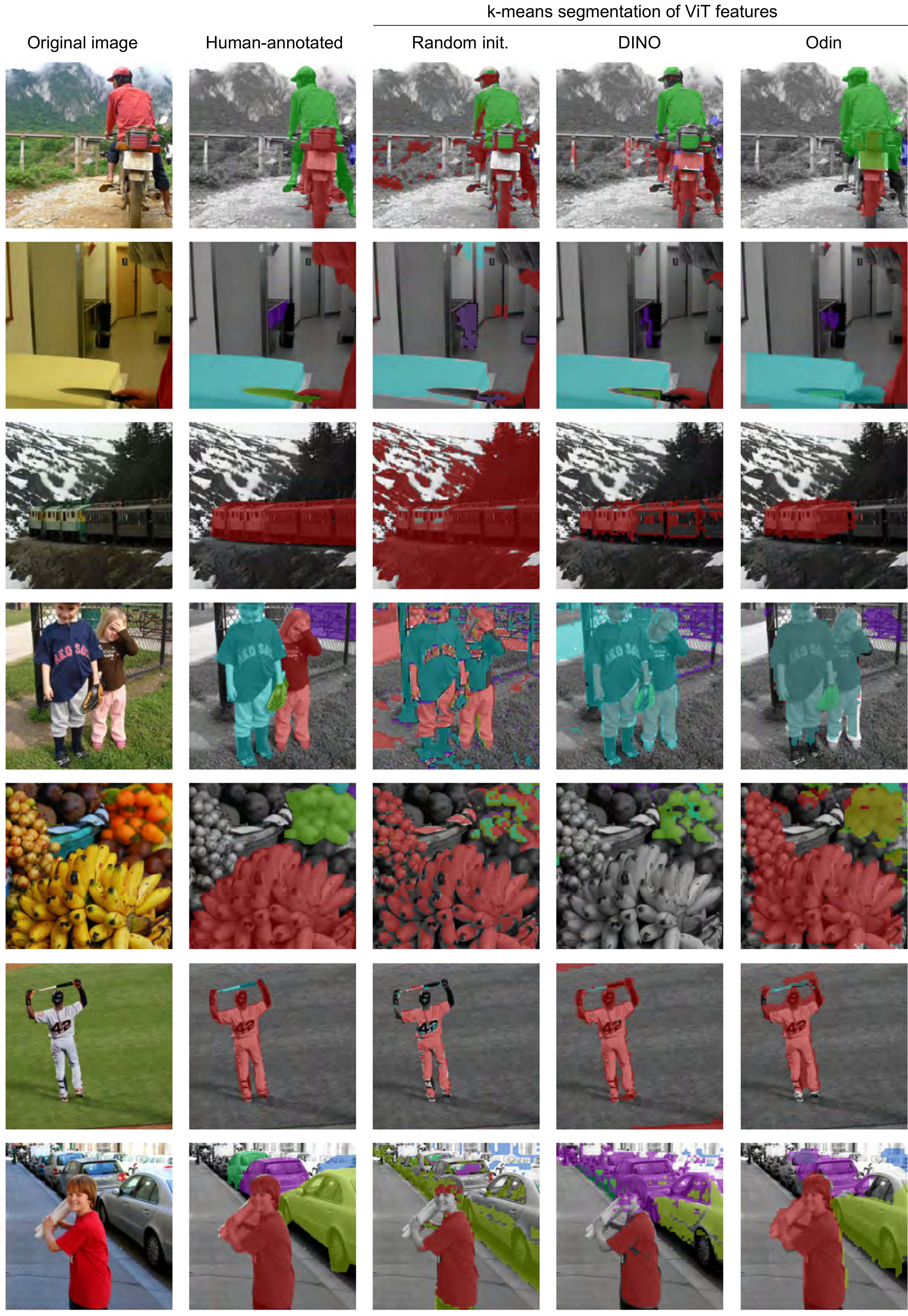}
\vspace{-2em}
\caption{Object discovery with vision transformers. \nth{1} column: original image, \nth{2}: human-annotated COCO segmentations, \nth{3}, \nth{4}, \nth{5}: segmentations obtained from k-means clustering on randomly intialized, DINO-, and Odin-trained vision transformer features, respectively.}
\label{fig:segmentations-vit}
\end{figure}

\subsection{Implementation: computational complexity}
\label{sec:app-complexity}

When continuously updating the discovery network, Odin requires an additional forward pass relative to the usual 2-view contrastive setup. This results in a +16\% computational overhead when using the same resolution as the other two views, or +67\% when doubling the image resolution. 
The k-means clustering of object-discovery features requires 0.25B FLOPS for standard-resolution images, or 1B FLOPS when doubling the resolution. When using a ResNet-50 backbone, these costs represent a +1.2\% and +5\% overhead respectively. 

When using a sparse set of discrete updates, e.g.\ every 100 epochs as in Table \ref{tab:ablation}, the unsupervised segmentations can be cached in-between updates. This reduces the  computational overhead by a factor of 100\x, making it negligible.

\subsection{Results: using object discovery or target networks for transfer learning}
\label{sec:app-target-teacher}

Although the object discovery and target networks are designed to provide a learning signal for the online network, we asked whether they could also be used for transfer learning. We therefore fine-tuned these networks for object detection on COCO and semantic segmentation on PASCAL and Cityscapes as before. We found their performance to be similar, but slightly worse than that of the online representation network (Table \ref{tab:app-target-teacher}). This is reasonable: once the online network has converged, its exponential moving average will largely catch up with it. The fact that they slightly underperform justifies our usage of the representation network for transfer.

\begin{table}[h]
\caption{\textbf{Transfer learning with different model parameters:} fine-tuning on COCO object detection and instance segmentation, and fine-tuning on PASCAL VOC and Cityscapes semantic segmentation are the same as in Table \ref{tab:prior_art_r50} (left) and Table \ref{tab:prior_art_seg}.}
\vspace{-.5em}
\small
\centering
\begin{tabularx}{0.8\linewidth}{c *{6}{Y}}
 \multicolumn{1}{c}{}
 & \multicolumn{2}{c}{COCO}
 & VOC & Citysc.
 \\
\cmidrule(lr){2-3} \cmidrule(l){4-5}
Model parameters & \apbbox{~} & \apmask{~} & \multicolumn{2}{c}{mIoU} \\
\midrule
target: $\xi$ & 42.7 & 38.3 & 78.5 & 76.6  \\ 
teacher: $\tau$ & 42.8 & \textbf{38.4} & 78.4 & 76.9  \\ 
online: $\theta$ & \textbf{42.9} & \textbf{38.4} & \textbf{78.6} & \textbf{77.1}  \\ 
\end{tabularx}
\label{tab:app-target-teacher}
\end{table}


\end{document}